\documentclass[10pt,twocolumn,letterpaper]{article}

\usepackage[pagenumbers]{cvpr} 
\usepackage{floatrow}
\usepackage{makecell}
\usepackage{multirow}
\usepackage{mathtools, cases}

\definecolor{cvprblue}{rgb}{0.21,0.49,0.74}
\usepackage[pagebackref,breaklinks,colorlinks,allcolors=cvprblue]{hyperref}

\title{ADV2E: Bridging the Gap Between Analogue Circuit and Discrete Frames \protect\\ in the Video-to-Events Simulator}

\author{Xiao Jiang\textsuperscript{1}  \quad Fei Zhou\textsuperscript{1} \quad Jiongzhi Lin\textsuperscript{1} \\
	\textsuperscript{1} Shenzhen University}

\begin{document}
\maketitle
\begin{abstract}
Event cameras operate fundamentally differently from traditional Active Pixel Sensor (APS) cameras, offering significant advantages. Recent research has developed simulators to convert video frames into events, addressing the shortage of real event datasets. Current simulators primarily focus on the logical behavior of event cameras. However, the fundamental analogue properties of pixel circuits are seldom considered in simulator design. The gap between analogue pixel circuit and discrete video frames causes the degeneration of synthetic events, particularly in high-contrast scenes. In this paper, we propose a novel method of generating reliable event data based on a detailed analysis of the pixel circuitry in event cameras. We incorporate the analogue properties of event camera pixel circuits into the simulator design: (1) analogue filtering of signals from light intensity to events, and (2) a cutoff frequency that is independent of video frame rate. Experimental results on two relevant tasks, including semantic segmentation and image reconstruction, validate the reliability of simulated event data, even in high-contrast scenes. This demonstrates that deep neural networks exhibit strong generalization from simulated to real event data, confirming that the synthetic events generated by the proposed method are both realistic and well-suited for effective training.
\end{abstract}
\section{Introduction}
\label{sec:intro}

Event cameras, e.g., Dynamic Vision Sensor (DVS) \cite{Delbruck1994, Lichtsteiner2008}, operate in a fundamentally different way than traditional cameras with Active Pixel Sensor (APS) \cite{Nakamura2006}. Unlike APS, which records scenes in discrete frames at fixed intervals, DVS continuously measures changes in light intensity asynchronously, outputting a stream of events. The outstanding properties of DVS, high dynamic range, low latency, and adaptive power consumption, offer significant advantages for numerous applications \cite{Alonso2019, Rebecq2021, Pan2019, Mueggler2017, Zhu2018, Gallego2022, Perot2020, Kim2008, Gehrig2021, Stoffregen2020} where traditional cameras are limited. 

As an infantile sensor, DVS suffers from scarcity and high expense \cite{Muglikar2021}. The event cameras currently on sale are highly specialized, requiring expert knowledge to carefully adjust their settings for real-world applications. The limited availability of event-based datasets for various vision tasks, such as semantic segmentation, is a technical constraint on the development of event camera. The conventional APS camera technology is highly developed, generating extensive datasets across a wide range of applications. This drives researchers to develop the simulators that can generate event data from existing APS datasets \cite{Kaiser2016, Koenig2004, Li2018, Rebecq2018, Gehrig2020, Hu2021, Lin2022, Zhang2024}. 

Recent methods concentrate on the logical behavior of event cameras, specifically by modeling the event generation process based on changes in light intensity between adjacent video frames. However, minimal attention has been given to the physical behaviors of DVS circuits, the fundamental analogue properties of pixel circuits in event cameras. In DVS, changes in light intensity are initially converted into electrical signals. These signals pass through multiple analogue components, ultimately producing a voltage difference that directly triggers event generation. The existing simulators overlook the analogue behaviors within the pixel circuits of DVS, leading to a degradation in the quality of synthetic events. The analogue behaviors are primarily characterized by a low-pass filtering process, with varying cutoff frequencies based on light intensity. Consequently, designing simulators without accounting for analogue behaviors results in inaccurate event timestamps, particularly in high-contrast scenes. 

This paper addresses the challenges involved in designing an event simulator. We conduct a detailed analysis of the DVS pixel circuit and find that the low-pass filters in analogue components, i.e., the cascode feedback loop and source follower, determine the time delay of event timestamps. We then introduce ADV2E, a novel event simulator, which incorporates the fundamental analogue behaviors of DVS. These behaviors include: (1) over-sampling brightness changes to ensure that the analogue filter's behaviors remain unaffected by frame rate variations, and (2) analog signal filtering through impulse invariance to preserve the accurate filtering characteristics of DVS. By embedding fundamental analogue properties, the proposed simulator markedly enhances performance in key vision tasks, such as semantic segmentation and image reconstruction, by generating realistic event data that is well-suited for training.

Our contributions are summarized as follows:

\begin{itemize}[leftmargin=!,labelindent=5pt]
	\item[$\bullet$] Our findings reveal that the disparity between existing event simulators and real DVS systems stems from the fundamental analogue behaviors within the pixel circuits of event cameras, particularly the low-pass filters in analogue circuit components.
	\item[$\bullet$] We propose a novel event simulator, ADV2E, which incorporates the behaviors of analogue circuit components in DVS to generate realistic event data from existing APS video frame datasets.
	\item[$\bullet$] We evaluate the proposed simulator both quantitatively and qualitatively, with a focus on time delay in high-contrast scenes, demonstrating its effectiveness compared to existing simulators.
	\item[$\bullet$] We validate the proposed simulator by generating synthetic events to train deep neural networks (DNNs) on common vision tasks, such as semantic segmentation and image reconstruction. Experimental results show that DNNs trained on simulated event data generalize well to real-world data.
\end{itemize}

\section{Related Work}
\label{sec:formatting}

\subsection{Event Cameras and Datasets}
\label{subsec:event_cameras_and_datasets}

Unlike APS cameras, event cameras asynchronously track brightness changes at each pixel. The output of event cameras consists of data streams, where each event is represented as $e_{i}=(x_{i}, y_{i}, t_{i}, p_{i})$. Here $i$ denotes the event index sorted by time, $(x_{i}, y_{i})$ represents the pixel location, $t_i$ is the timestamp, and $p_{i}\in\{-1,+1\}$ indicates the polarity, signifying a negative or positive brightness change, respectively. The logical behavior of event cameras is defined by the condition ${|L(x_i, y_i, t_i) - L(x_i, y_i, t_i-t_p)|\geq C}$, where $L(\cdot)$ denotes the logarithmic operation, $t_{p}$ is the timestamp of $e_p$, the last triggered event before $t_{i}$, and $C$ is the predefined threshold of DVS. This means that an event is triggered when the change in light intensity exceeds the threshold $C$ since $e_p$ is generated.

In reality, the DVS circuit operates in a complex manner. Changes in the light signal pass through multiple stages of analog components, including the cascode feedback loop and source follower, before events are generated. These components are designed to amplify and stabilize signals, yet their fundamental analogue properties introduce additional, intricate behaviors. Recent research \cite{Graca2023, Graca2023_1, McReynolds2023} suggests that these behaviors can be approximated as a low-pass filter, though the exact order of the filter remains unclear.

As a burgeoning technology, event cameras are produced by only a handful of specialized companies\footnote{https://inivation.com/buy/; https://www.shop.prophesee.ai} with a high price tag. The real event-based datasets are either strongly limited by scenes \cite{Hu2020, Kim2021, Luo2023, Tournemire2020} (e.g., fixed patterns captured in indoors) or composed of sequences with brief duration \cite{Mueggler2017}. Datasets for high-level vision tasks, \textit{e.g.} semantic segmentation, are still scarce due to the difficulty of labeling objects within event streams \cite{Alonso2019}.

\subsection{Event Simulators}
Kaiser \textit{et al.}\ \cite{Kaiser2016} introduce a basic simulator that directly measures brightness changes between adjacent frames, outputting events synchronously. In standard videos with frame rates of 24–30 frames per second (FPS), this method results in simulated event timestamps deviating by up to tens of milliseconds. Rebecq \textit{et al.}\ \cite{Rebecq2018} and Gehrig \textit{et al.}\ \cite{Gehrig2020} address this issue with an adaptive sampling strategy, interpolating video frames non-uniformly to ensure asynchronous event timestamps, attributing imprecise event timing to the synchronous nature of video frames. Hu \textit{et al.}\ \cite{Hu2021}  analyze the DVS pixel circuit and identify event blur due to the low-pass filter effect in real event cameras, which introduces latency in event generation. In typical DVS configurations, this latency occurs on the millisecond scale, suggesting that the fundamental analogue properties of DVS circuits—specifically, the low-pass filter—are a primary source of timestamp imprecision when frame intervals span milliseconds. Consequently, Hu \textit{et al.}\ \cite{Hu2021} employ optical flow \cite{Jiang2018} to interpolate frames uniformly from tens to hundreds of FPS, incorporating a basic low-pass filter in their v2e simulator to enhance timestamp accuracy. Additionally, v2e models noise in DVS circuits, such as temporal and leakage noise; however, its low-pass filter discretizes analogue properties in an overly simplistic manner, overlooking the continuity of signal changes in DVS circuits, which affects timestamp accuracy.  Lin \textit{et al.}\ \cite{Lin2022} propose an event generation model based on Brownian motion with drift, assuming a linear brightness change that does not align with analog filter behavior, leading to timestamp inaccuracies. This simulator requires the calibration of six parameters per scene, with some parameters needing manual adjustment. Zhang \textit{et al.}\ \cite{Zhang2024} utilize a neural network to generate time bins for events, subsequently deriving event timestamps using a probability model. This model assumes a uniform distribution of events within each time bin, applying a linear probability density function. However, this assumption does not hold in the presence of low-pass filter behaviors in real DVS circuits. In summary, the existing research either disregards the low-pass filter effect or incorporates it in an excessively simplistic manner, overlooking the continuous nature of brightness changes.

\section{ADV2E Simulator}

\subsection{Problem Formulation}
\label{subsec:Problem_formulation}

Let $\{I_n\}^{N-1}_{n=0}$ represent a sequence of consecutive APS frames in a video. Typically, the interval $T_b$ between adjacent frames remains constant and corresponds to the reciprocal of the frame rate $f_b$. The objective of the event simulator is to generate realistic event streams $\{e_i\}$, where the timestamp $t_i$ spans from $0$ to $T$.

\begin{figure*}
	\centering
	\includegraphics[width=16.5cm]{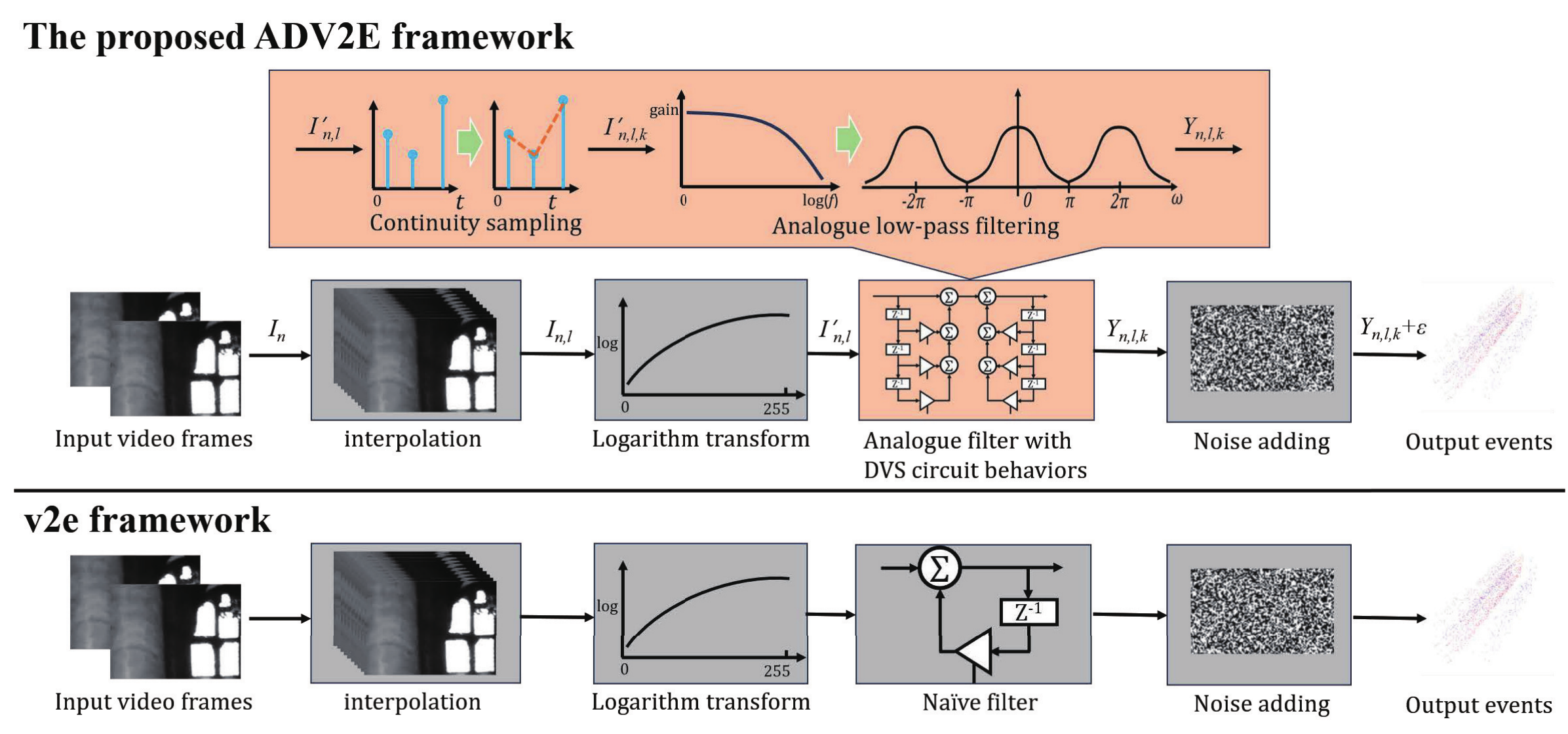}
	\vspace{-3mm}
	\caption{The framework of the proposed ADV2E simulator. Different from v2e, the proposed simulator models brightness changes by faithfully emulating the core analogue behaviors inherent to real DVS circuitry, including continuity sampling and analogue low-pass filtering.  Therefore, the proposed event simulator addresses the issue of inaccurate filtering delays in current simulators, improving the accuracy of simulated events.}
	\label{fig:framework}
\end{figure*}

We adopt the similar framework as v2e for event generation. However, unlike v2e, the proposed ADV2E simulator processes brightness changes by closely replicating the core analogue behaviors of real DVS, as illustrated in Fig. \ref{fig:framework} and detailed below. The other components, including interpolation, logarithmic transformation, and noise addition, are shared between the two simulators. For clarity, we denote the outputs of interpolation as $\{I_{n,l}\}_{n=0,l=0}^{N-1,L-1}$, where $L$ is the interpolation factor, and the results of the logarithmic transformation as $\{I^{\prime}_{n,l}\}_{n=0,l=0}^{N-1,L-1}$.

\subsection{Method}
\label{subsec:Method}

In this section, we present a detailed overview of the proposed simulator, as highlighted in orange in Fig. \ref{fig:framework}. As demonstrated in the supplementary materials, the core analogue behaviors are effectively modeled as a first-order low-pass filter. Our primary focus is on the enhancements applied to this low-pass filter. The filter component takes the output of the logarithmic transformation as input, producing results used for noise addition to derive events directly. The low-pass filter enhancements comprise two primary components: (1) continuity sampling, which prevents aliasing and preserves essential analogue behaviors; (2) an analog filter designed with impulse invariance, enabling precise filtering of discrete video frames.

\subsubsection{Continuity Sampling}

\begin{figure}
	\centering
	\includegraphics[width=0.8\linewidth]{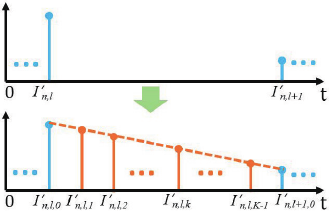}	
	\caption{Continuity Sampling. The frame rate is significantly increased through linear over-sampling, effectively preventing aliasing and preserving the analogue behaviors of DVS.}
	\label{fig:continuity_sampling}
\end{figure}

We over-sample the signals linearly in Fig. \ref{fig:continuity_sampling}, effectively preventing aliasing and enabling a continuous low-pass filter that dynamically adjusts to brightness levels. While APS video frames are inherently discrete, DVS functions in an analogue manner. Thus, the event simulator must discretize the analog behaviors of DVS to produce events. As discussed in \cite{Lichtsteiner2008} and the supplemental material, low-pass filtering is a fundamental aspect of DVS circuits. The latency for nominal mode is approximately 4 ms \cite{delbruck2020}, corresponding to a cutoff frequency of around 250 Hz . According to the Nyquist–Shannon sampling theorem \cite{Shannon1949}, the discrete signal should be sampled at a rate at least twice the signal bandwidth to prevent aliasing. This implies that the minimum frame rate for APS video frames, even in nominal DVS mode, should be 500 FPS. 
The videos require an inter-frame interpolation factor $L$ of at least 16x. However, existing video frame interpolation techniques are not capable of supporting such a high interpolation factor \cite{Dong2023}.

To tackle this challenge, the frames $\{I^{\prime}_{n,l}\}_{n=0,l=0}^{N-1,L-1}$ are linearly over-sampled by a factor of $K$, as illustrated below:

\begin{equation}
	I^{\prime}_{n,l,k} = 
	\begin{cases}
		I^{\prime}_{n,l} + \frac{I^{\prime}_{n,l+1}-I^{\prime}_{n,l}}{K}\cdot k & if ~ l \neq L-1, \\
		I^{\prime}_{n+1,0} + \frac{I^{\prime}_{n+1,0}-I^{\prime}_{n,L-1}}{K}\cdot k & if ~ l = L-1,
	\end{cases}
	\label{eq:continuity_sampling}
\end{equation}

\noindent where $\{I^{\prime}_{n,l,k}\}$ represent the sampled frames, with $k$ denoting the number of over-sampled frames. Since the time interval $\frac{Tb}{N\cdot L}$ is brief, the brightness change can be approximated as linear. Through continuity sampling, the effective sampling rate increases from $L\cdot f_b$ to $K\cdot L \cdot f_b$. Consequently, the aliasing caused by limited frame interpolation can be effectively mitigated by setting $K$ such that $K\cdot L \cdot f_b$ exceeds the Nyquist frequency, without regard to the inter-frame interpolation factor $L$ or the original frame rate $f_b$.

\subsubsection{Analogue Low-pass Filtering}

\begin{figure}[t]
	\centering
	\includegraphics[width=0.8\linewidth]{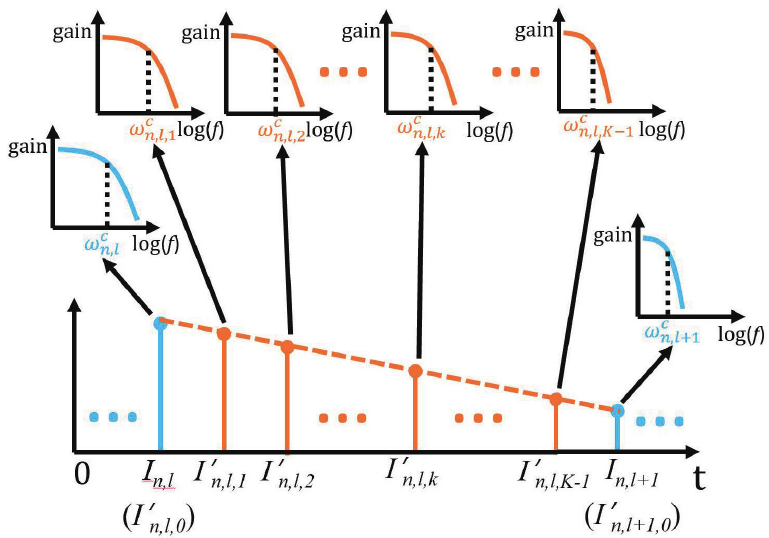}
	\vspace{-2mm}
	\caption{Analogue low-pass filtering. In a real DVS circuit, the cutoff frequency of the low-pass filters varies over time, proportional to the current brightness. By applying continuous brightness values, the proposed simulator closely approximates the behaviors of DVS analogue circuit, enabling realistic event generation. }
	\label{fig:analogue_filtering}
\end{figure}

The fundamental analogue behaviors of the DVS circuit can be summarized as a first-order low-pass filter, as detailed in the supplemental material. The system function is:

\begin{equation}
	H_a(s)=\frac{\omega_{0}(t)}{s+\omega_{0}(t)}
	\label{eq:analog_lowpass_filter}
	,\end{equation}
	
\noindent where $\omega_{0}(t)$ is the time-varying cutoff frequency of the low-pass filter. The cutoff frequency $\omega_{0}(t)$ is directly proportional to the current brightness \cite{delbruck2020}, $\omega_{0}(t)\propto I(t)$, where $I(t)$ represents the brightness intensity at time $t$. With the proposed continuity sampling, light intensity variations are minimized and can be assumed constant within the brief interval $\frac{T_b}{KL}$, as the over-sampling rate K can theoretically be increased indefinitely. Assuming that:
\begin{equation}
	\omega_0^{n,l,k}=\omega_0((n+\frac{l}{L}+\frac{k}{KL})T_b), 
	\label{eq:frequency_division}
\end{equation}
 
\noindent which is the cutoff frequency proportional to $I_{n,l,k}$ at the same moment, the proposed ADV2E provides the low-pass filtered result as follows:

\begin{equation}
	Y = e^{-\alpha} \cdot Y + \alpha \cdot I^{\prime}_{n,l,k}, 
\label{eq:digital_lowpass_filter}
\end{equation}

\noindent where $\alpha=\frac{\omega_0^{n,l,k}T_b}{KL}$ is the filter coefficient. Compared to vide2e \cite{Rebecq2018, Gehrig2020}, DVS-Voltmeter \cite{Lin2022}, and v2ce \cite{Zhang2024}, the proposed ADV2E integrates the fundamental analogue behaviors of DVS circuitry, specifically the low-pass filter, into the event simulation process, thereby significantly enhancing event generation accuracy. In the case of v2e \cite{Hu2021}, the interpolation factor $L$ is constrained by current frame interpolation techniques, which can lead to aliasing of the low-pass filter. The time interval between adjacent interpolated frames, $\frac{T_b}{L}$, is insufficiently small to approximate stable light intensity, causing inaccuracies in the low-pass filter, as the cutoff frequency varies with light intensity. Unlike v2e, ADV2E utilizes continuity sampling to substantially reduce the time interval, effectively eliminating low-pass filter aliasing. This approach further improves low-pass filter accuracy by maintaining a constant cutoff frequency over the smaller interval $\frac{T_b}{KL}$. In summary, by bridging the gap between the analog pixel circuitry and discrete video frames, ADV2E greatly enhances the realism of generated events, closely resembling real event data.

\section{Experimental Results}

\subsection{Implementation Details}

\noindent \textbf{Settings.} In the proposed ADV2E simulator, the continuity sampling rate $K$ is the only parameter requiring adjustment. Here, we set $K$ to $10$. All other parameters match those in v2e and vid2e, where the frame interpolation rate $L$ is set to $10$. With these settings, aliasing of the low-pass filter in the DVS circuit is effectively avoided. Since the frame rate of typical videos is at least 24 FPS, the corresponding sampling frequency with continuity sampling is ${KLf_b=2400Hz}$ by continuity sampling. This rate exceeds greatly the cutoff frequency of the low-pass filter in DVS circuits operating in the commonly used nominal mode. We present comprehensive comparisons between the ADV2E simulator and four state-of-the-art (SOTA) event simulators, including vid2e \cite{Rebecq2018, Gehrig2020}, v2e \cite{Hu2021}, DVS-Voltmeter \cite{Lin2022}, and v2ce \cite{Zhang2024}.

\noindent \textbf{Tasks.} Three tasks are conducted to evaluate the proposed ADV2E and other simulators. First, we directly compare the event simulators by visualizing the generated events. The compared methods are further evaluated on two tasks: semantic segmentation and image reconstruction.

\begin{itemize}
	\item[$\bullet$] \textbf{Direct Comparisons: } The 'urban' sequence from the DAVIS240C dataset \cite{Mueggler2017} is used for direct comparisons of generated events across different simulators. This dataset provides APS video frames with timestamps alongside corresponding ground truth events. The 'urban' sequence is selected for its high-contrast scenes featuring diverse light intensities and dynamic motion, which allows for a clear analysis of the low-pass filter effects, as its cutoff frequency varies in proportion to light intensity. The simulated events are evaluated both qualitatively and quantitatively. Events are displayed in standard format \cite{Berner2007}. Due to the lack of pixel-level quantitative comparisons between events, we use the evaluation based on the commonly adopted representation of temporal bins \cite{Rebecq2021}. We represent the events using temporal bins and measure the Euclidean distance between the simulated and real events.
	\item[$\bullet$] \textbf{Semantic Segmentation:} We use the DDD17 dataset \cite{Binas2017}, an open driving dataset captured by event cameras, for comparisons. Semantic annotations for training and testing are provided by Ev-Segnet \cite{Alonso2019}, with six class labels: 'flat,' 'construction+sky,' 'object,' 'nature,' 'human,' and 'vehicle.' Events generated by each simulator are separately used to train the segmentation network Ev-Segnet from scratch to assess its generalization ability on real test data. Training setups also adhere to the original recommendations. For evaluation, we employ standard metrics, including accuracy and mIoU.
	\item[$\bullet$] \textbf{Image Reconstruction: } We select the E2VID network \cite{Rebecq2021} as the image reconstruction pipeline. E2VID takes events as input and outputs reconstructed APS frames. Training data is synthesized by event simulators from the GOPRO dataset \cite{Nah2017}, which contains 22 high-quality video frame sequences. As with E2VID and DVS-Voltmeter, we use seven sequences from the DAVIS240C dataset for validation on real test data. These sequences include 'dynamic{\textunderscore}6dof,' 'boxes{\textunderscore}6dof,' 'poster{\textunderscore}6dof,' 'shapes{\textunderscore}6dof,' 'office{\textunderscore}zigzag,' 'slider{\textunderscore}depth,' and 'calibration.' For quantitative evaluation of reconstructed image quality, we employ MSE, SSIM \cite{Zhou2004}, and LPIPS \cite{Zhang2018}. 
\end{itemize}

\subsection{Direct Comparisons}

\begin{figure}
	\centering
	\begin{subfigure}[b]{\linewidth}
		\includegraphics[width=\linewidth]{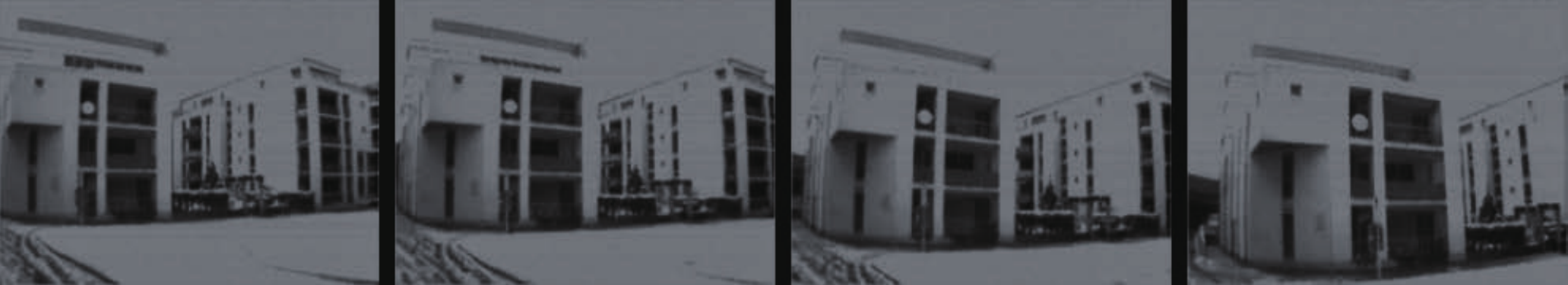}
		\caption{APS frames.}
		\label{fig:realEventsAPS}
	\end{subfigure}
	\vfill
	\begin{subfigure}[b]{\linewidth}
		\includegraphics[width=\linewidth]{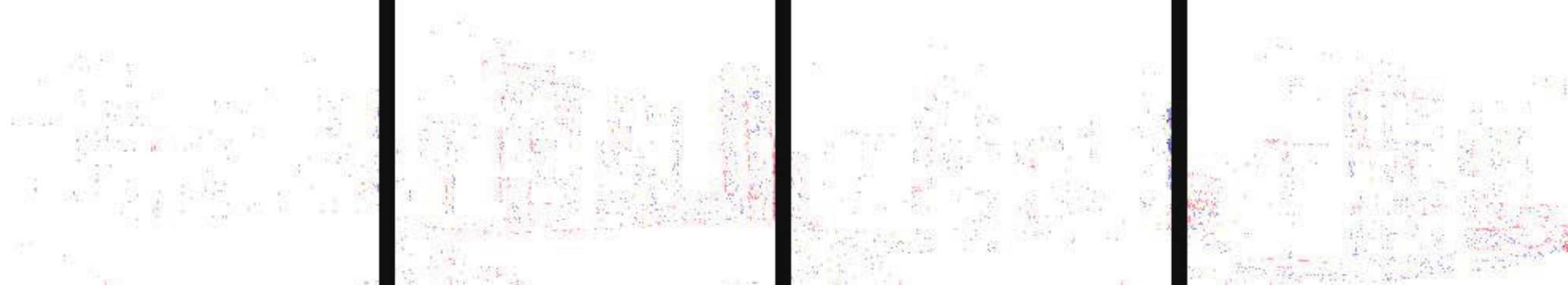}
		\caption{vid2e \cite{Rebecq2018, Gehrig2020}.}
		\label{fig:realEventsVid2e}
	\end{subfigure}
	\begin{subfigure}[b]{\linewidth}
		\includegraphics[width=\linewidth]{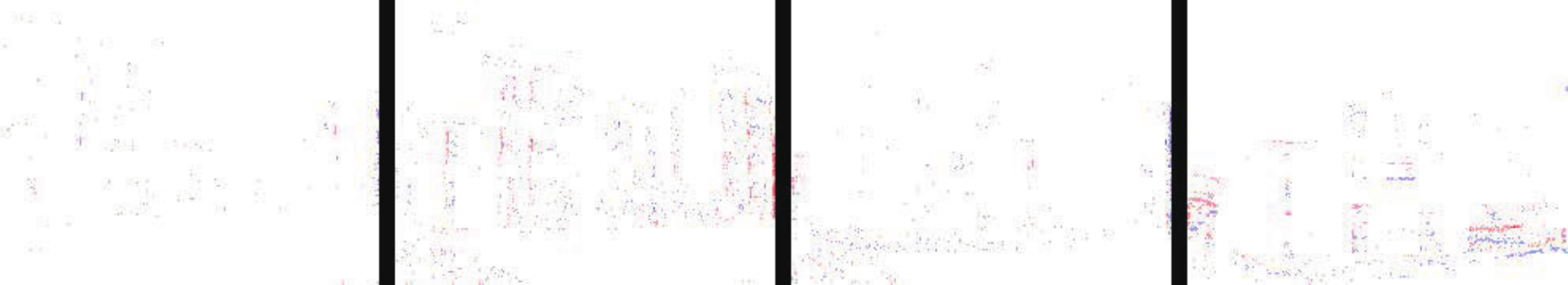}
		\caption{v2e \cite{Hu2021}.}
		\label{fig:realEventsV2e}
	\end{subfigure}
	\begin{subfigure}[b]{\linewidth}
		\includegraphics[width=\linewidth]{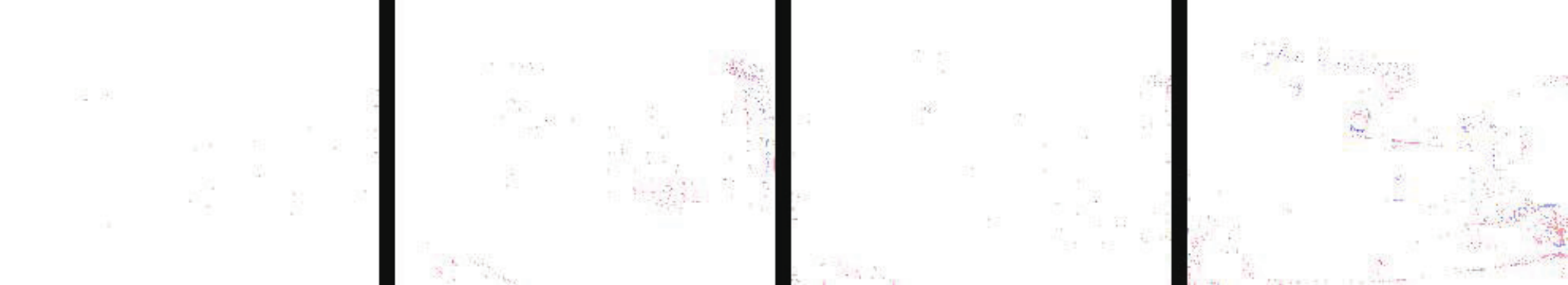}
		\caption{v2ce \cite{Zhang2024}.}
		\label{fig:realEventsV2ce}
	\end{subfigure}
	\begin{subfigure}[b]{\linewidth}
		\includegraphics[width=\linewidth]{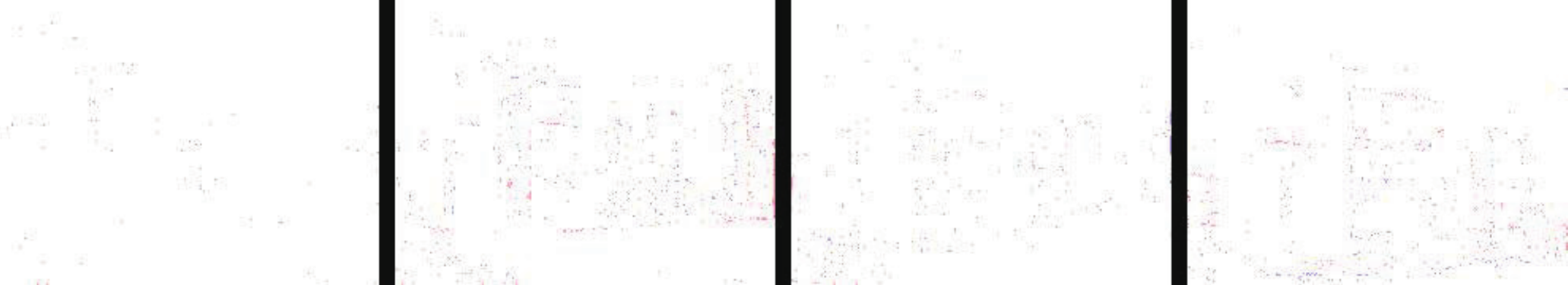}
		\caption{DVS-Voltmeter \cite{Lin2022}.}
		\label{fig:realEventsDVSVoltmeter}
	\end{subfigure}
	\begin{subfigure}[b]{\linewidth}
		\includegraphics[width=\linewidth]{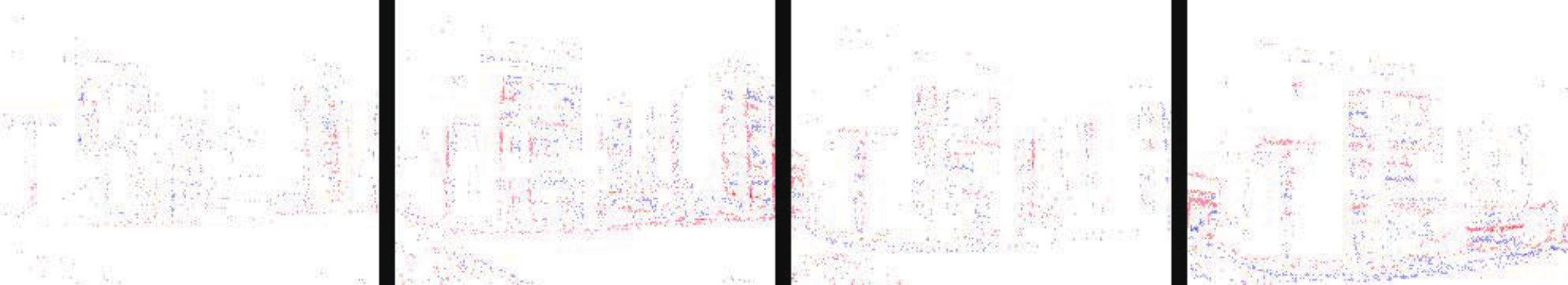}
		\caption{the proposed ADV2E.}
		\label{fig:realEventsADV2E}
	\end{subfigure}
	\begin{subfigure}[b]{\linewidth}
		\includegraphics[width=\linewidth]{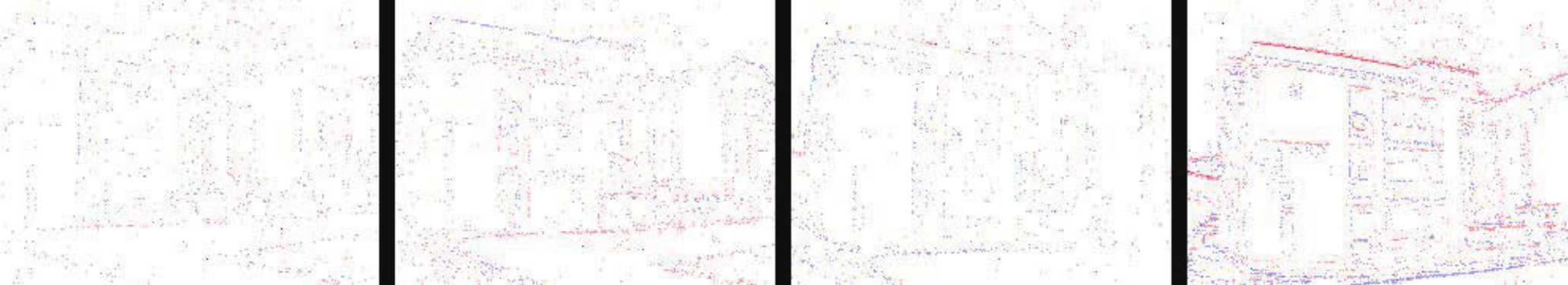}
		\caption{ground truth.}
		\label{fig:realEventsGT}
	\end{subfigure}
	\vspace{-6mm}
	\caption{Qualitative comparison of events generated by different simulators. Details are best viewed when zoomed in. Each column represents a frame from the DAVIS dataset \cite{Mueggler2017}. From top to bottom, there are  (a) APS frames, events synthesized by (b) vid2e \cite{Rebecq2018, Gehrig2020}, (c) v2e \cite{Hu2021}, (d) v2ce \cite{Zhang2024}, (e) DVS-Voltmeter \cite{Lin2022}, (f) the proposed ADV2E, and (g) ground truth. The events produced by ADV2E most closely resemble the ground truth, with realistic event generation in high-contrast regions, such as the balconies and the intersections between buildings and roads.}
	\label{fig:realEvents}
\end{figure}

\begin{table}
	\centering
	\resizebox{\textwidth}{!}
	{
	\begin{tabular}{cccccc}
		\toprule
		  & vid2e \cite{Rebecq2018, Gehrig2020} & v2e \cite{Hu2021} & v2ce \cite{Zhang2024} & \makecell{DVS-\\Voltmeter \cite{Lin2022}} & ADV2E \\
		\midrule
		distance $\downarrow$  & 61.01 & 68.64 & \textit{51.97} & 71.49 & \textbf{49.58} \\
		\bottomrule
	\end{tabular}
	}
	\vspace{-6mm}
	\caption{Euclidean distance between the temporal bins \cite{Rebecq2021} of simulated and real events. Here a smaller value indicates greater similarity. As can be seen, the events synthesized by the proposed ADV2E is the most similar to the real events. The \textbf{bold} and \textit{italic} refer to the best and second best, respectively.}
	\label{tab:realEvents}
\end{table}

The events generated by the proposed ADV2E and other simulators are compared in Fig. \ref{fig:realEvents} and Table \ref{tab:realEvents}. The APS frames reveal dramatic contrast between light and shadow, with intense changes in relative speed between the camera and the scene. These frames capture moments when the relative velocity suddenly decreases. Without accounting for the fundamental analogue behaviors—specifically, the low-pass filter within the DVS circuit’s analog components—the generated events would abruptly vanish, as seen in the results from vid2e, v2ce, and DVS-Voltmeter. However, the ground truth shows that events persist due to the time delay introduced by the DVS circuit’s low-pass filter, which causes event generation to continue for a duration depending on the filter's cutoff frequency.

The v2e simulator incorporates a basic low-pass filter to account for this delay, but it lacks continuity in analog signal processing, leading to aliasing and inaccuracies in the filter’s cutoff frequency. In contrast, the proposed ADV2E simulator employs continuity sampling with a time-dependent cutoff frequency, enhancing the accuracy of the low-pass filter. As shown in Fig. 2, ADV2E generates events that are more realistic than those from v2e. By carefully embedding the fundamental analogue behaviors of the DVS circuit, ADV2E achieves the most realistic event generation among all tested simulators.

Table \ref{tab:realEvents} provides quantitative comparisons between the event simulators. The smallest distance between temporal bins indicates that ADV2E achieves the highest similarity to real events, validating the effectiveness of the fundamental analog behaviors incorporated by the proposed method.

\subsection{Semantic Segmentation}

\begin{figure*}
	\centering
	\begin{subfigure}[b]{\linewidth}
		\includegraphics[width=\linewidth]{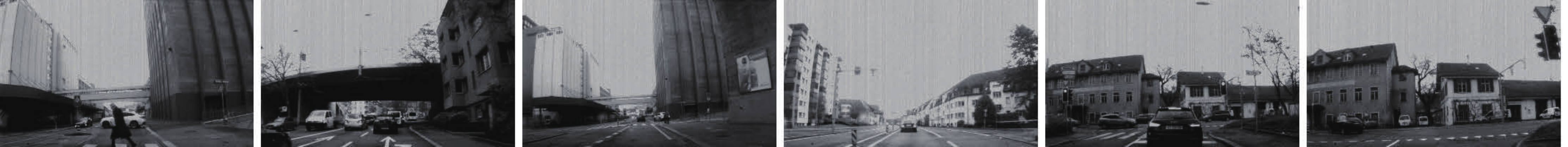}
		\caption{APS frames.}
		\label{fig:EvSegnetAPS}
	\end{subfigure}
	\vfill
	\begin{subfigure}[b]{\linewidth}
		\includegraphics[width=\linewidth]{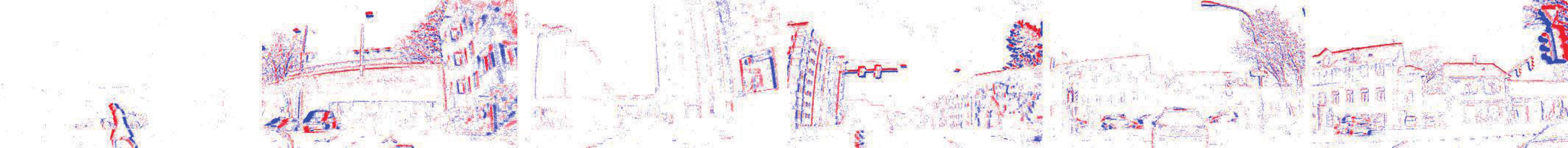}
		\caption{events.}
		\label{fig:EvSegnetEvents}
	\end{subfigure}
	\vfill
	\begin{subfigure}[b]{\linewidth}
		\includegraphics[width=\linewidth]{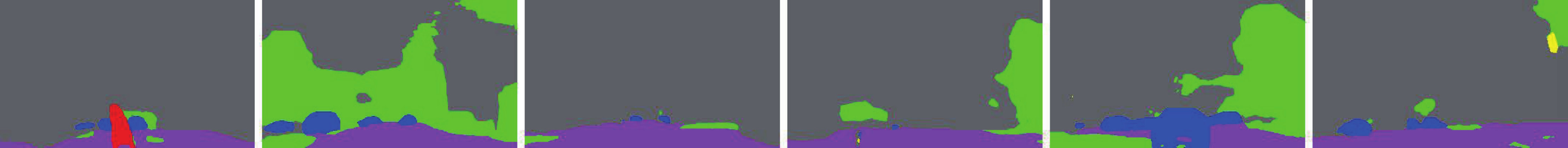}
		\caption{vid2e \cite{Rebecq2018, Gehrig2020}.}
		\label{fig:EvSegnetVid2e}
	\end{subfigure}
	\begin{subfigure}[b]{\linewidth}
		\includegraphics[width=\linewidth]{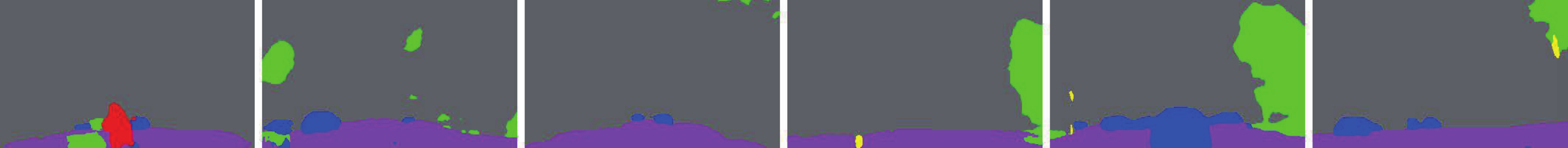}
		\caption{v2e \cite{Hu2021}.}
		\label{fig:EvSegnetV2e}
	\end{subfigure}
	\begin{subfigure}[b]{\linewidth}
		\includegraphics[width=\linewidth]{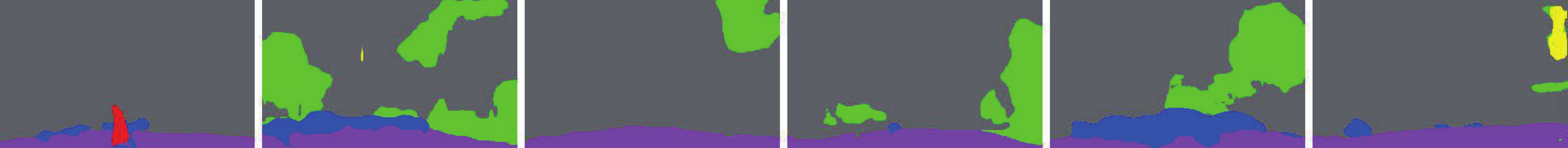}
		\caption{v2ce \cite{Zhang2024}.}
		\label{fig:EvSegnetV2ce}
	\end{subfigure}
	\begin{subfigure}[b]{\linewidth}
		\includegraphics[width=\linewidth]{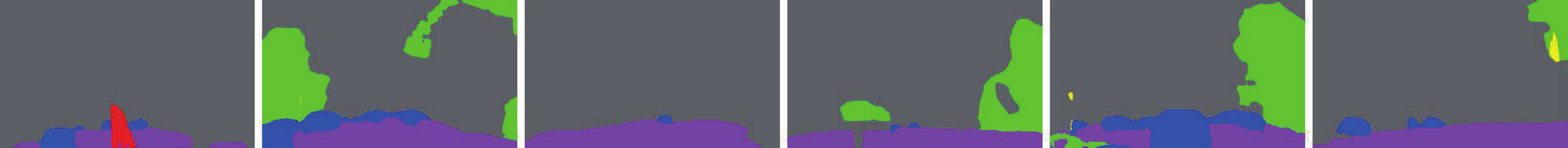}
		\caption{DVS-Voltmeter \cite{Lin2022}.}
		\label{fig:EvSegnetDVSVoltmeter}
	\end{subfigure}
	\begin{subfigure}[b]{\linewidth}
		\includegraphics[width=\linewidth]{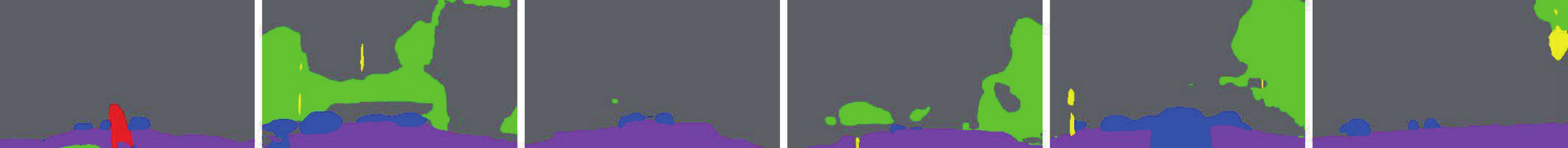}
		\caption{the proposed ADV2E.}
		\label{fig:EvSegnetADV2E}
	\end{subfigure}
	\begin{subfigure}[b]{\linewidth}
		\includegraphics[width=\linewidth]{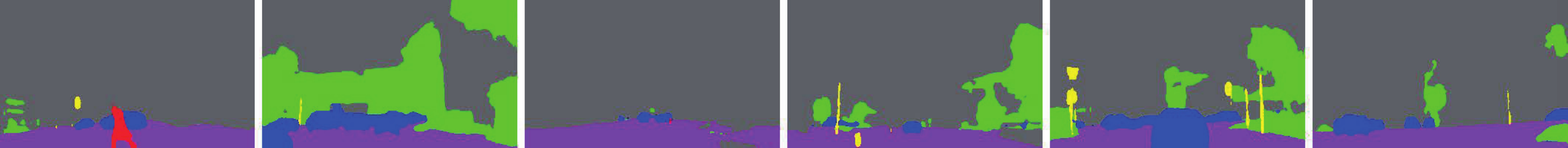}
		\caption{ground truth.}
		\label{fig:EvSegnetGT}
	\end{subfigure}
	\vspace{-6mm}
	\caption{ Qualitative comparison of semantic segmentation across event simulators. Each column represents a separate scene. From top to bottom, the rows show (a) APS frames, (b) events, segmentation results by (c) vid2e \cite{Rebecq2018, Gehrig2020}, (d) v2e \cite{Hu2021}, (e) v2ce \cite{Zhang2024}, (f) DVS-Voltmeter \cite{Lin2022}, (g) the proposed ADV2E, and (h) ground truth. The ground truth is generated automatically by a CNN \cite{Alonso2019} on APS images. Compared to other simulators, the events generated by the proposed ADV2E simulator enable the CNN to capture objects with greater accuracy. For example, the pedestrian in the 1st column, the bridge in the 2nd column, and all vehicles are segmented precisely by the proposed ADV2E. In the 5th column, only the network trained on events from ADV2E correctly identifies the traffic light on the left. Overall, ADV2E produces the most realistic events among the simulators evaluated.}
	\label{fig:EvSegnet}
\end{figure*}

\begin{table}
	\centering
	\resizebox{\textwidth}{!}
	{
		\begin{tabular}{ccccccc}
			\toprule
			& vid2e & v2e & v2ce & \makecell{DVS-\\Voltmeter} & ADV2E & Real \\
			\midrule
			accuracy $\uparrow$  & 84.95 & 84.11 & 86.61 & \textbf{87.88} & \textit{87.14} & 89.76\\
			mIoU $\uparrow$  & 46.67 & 42.25 & 48.84 & \textit{50.60} & \textbf{51.20} & 54.81\\
			\bottomrule
		\end{tabular}
	}
	\vspace{-5mm}
	\caption{Performance of semantic segmentation. Networks are trained on simulated events and then tested on real event data \cite{Binas2017}. The results trained by the real event data are also given below. The \textbf{bold} and \textit{italic} refer to the best and second best, respectively. The reason that DVS-Voltmeter has slightly higher accuracy than the proposed ADV2E is that DVS-Voltmeter tends to misclassify other objects as background, as can be seen in Fig. \ref{fig:EvSegnet}.}
	\label{tab:EvSegnet}
\end{table}

Qualitative and quantitative results are presented in Fig. \ref{fig:EvSegnet} and Table \ref{tab:EvSegnet}, respectively. In Fig. \ref{fig:EvSegnet}, the proposed ADV2E simulator produces events that enable the CNN to segment objects with high accuracy. For example, in the 1st column, the pedestrian is outlined with a sharp contour around the legs, a level of detail that other methods fail to achieve. Vehicles are clearly labeled across all frames, except for those at great distances, caused by limitation of DVS resolution. ADV2E also successfully captures traffic lights, as seen in the 5th column, which other methods miss. Additionally, ADV2E accurately segments background elements, such as the bridge in the 2nd column and road surfaces throughout.

Table \ref{tab:EvSegnet} shows that ADV2E achieves the highest mIoU, closely approaching the performance of networks trained on real event data. The accuracy achieved by ADV2E is the second highest, just below that of DVS-Voltmeter. This occurs because the DVS-Voltmeter tends to misclassify objects as background, and the background occupies the majority of the image area. These experimental results indicate that events generated by ADV2E are the most realistic among the evaluated simulators for semantic segmentation.

\subsection{Image Reconstruction}

\begin{table}
	\centering
	\resizebox{\textwidth}{!}
	{
		\begin{tabular}{cccccccc}
			\toprule
			& Metric & vid2e & v2e & v2ce & \makecell{DVS-\\Voltmeter} & ADV2E \\
			\midrule
			\multirow{3}{*}{dynamic\textunderscore 6dof} & MSE $\downarrow$ & 0.09 & 0.18 & 0.33 & \textit{0.05} & \textbf{0.04} \\
			& SSIM $\uparrow$ & 0.37 & 0.23 & 0.20 & \textbf{0.43} & \textit{0.41} \\
			& LPIPS $\downarrow$ & 0.37 & \textit{0.44} & 0.53 & \textbf{0.41} & 0.47 \\
			\midrule
			\multirow{3}{*}{boxes\textunderscore 6dof} & MSE $\downarrow$ & 0.04 & 0.12 & 0.06 & \textit{0.03} & \textbf{0.02} \\
			& SSIM $\uparrow$ & 0.51 & 0.28 & \textbf{0.53} & \textit{0.52} & \textbf{0.53} \\
			& LPIPS $\downarrow$ & \textit{0.47} & 0.59 & \textbf{0.42} & \textit{0.47} & \textbf{0.42} \\
			\midrule
			\multirow{3}{*}{poster\textunderscore 6dof} & MSE $\downarrow$ & 0.08 & 0.16 & 0.16 & \textit{0.04} & \textbf{0.01} \\
			& SSIM $\uparrow$ & 0.43 & 0.23 & 0.31 & \textit{0.50} & \textbf{0.56} \\
			& LPIPS $\downarrow$ & \textbf{0.35} & 0.51 & 0.44 & \textit{0.37} & 0.38 \\
			\midrule
			\multirow{3}{*}{shapes\textunderscore 6dof} & MSE $\downarrow$ & \textit{0.02} & 0.05 & 0.07 & \textbf{0.01} & \textbf{0.01} \\
			& SSIM $\uparrow$ & 0.71 & 0.63 & 0.62 & \textit{0.79} & \textbf{0.81} \\
			& LPIPS $\downarrow$ & \textit{0.35} & 0.38 & 0.56 & \textbf{0.28} & 0.48 \\
			\midrule
			\multirow{3}{*}{office\textunderscore zigzag} & MSE $\downarrow$ & 0.06 & 0.13 & 0.07 & 0.04 & \textbf{0.02} \\
			& SSIM $\uparrow$ & 0.43 & 0.23 & 0.44 & \textit{0.46} & \textbf{0.47} \\
			& LPIPS $\downarrow$ & 0.51 & 0.60 & \textbf{0.39} & 0.48 & \textit{0.42} \\
			\midrule
			\multirow{3}{*}{slider\textunderscore depth} & MSE $\downarrow$ & 0.05 & 0.11 & \textit{0.03} & \textit{0.03} & \textbf{0.02} \\
			& SSIM $\uparrow$ & 0.41 & 0.34 & \textit{0.52} & 0.46 & \textbf{0.54} \\
			& LPIPS $\downarrow$ & 0.52 & 0.56 & \textbf{0.45} & 0.50 & \textit{0.46} \\
			\midrule
			\multirow{3}{*}{calibration} & MSE $\downarrow$ & 0.05 & 0.12 & 0.06 & \textit{0.04} & \textbf{0.02} \\
			& SSIM $\uparrow$ & 0.54 & 0.39 & \textbf{0.58} & \textit{0.55} & \textit{0.55} \\
			& LPIPS $\downarrow$ & 0.47 & 0.55 & \textbf{0.41} & \textit{0.42} & 0.45 \\
			\midrule
			\multirow{3}{*}{average} & MSE $\downarrow$ & 0.06 & 0.12 & 0.22 & \textit{0.03} & \textbf{0.02} \\
			& SSIM $\uparrow$ & 0.51 & 0.35 & 0.46 & \textit{0.53} & \textbf{0.55} \\
			& LPIPS $\downarrow$ & 0.41 & 0.50 & 0.46 & \textbf{0.42} & \textit{0.44} \\
			\bottomrule
		\end{tabular}
	}
	\vspace{-5mm}
	\caption{Performance of image reconstruction. Networks are trained on simulated events  and then tested on real event data \cite{Mueggler2017}. The \textbf{bold} and \textit{italic} refer to the best and second best, respectively.}
	\label{tab:eventsCnnMinimal}
\end{table}

\begin{figure*}
	\centering
	\begin{subfigure}[b]{\linewidth}
		\includegraphics[width=\linewidth]{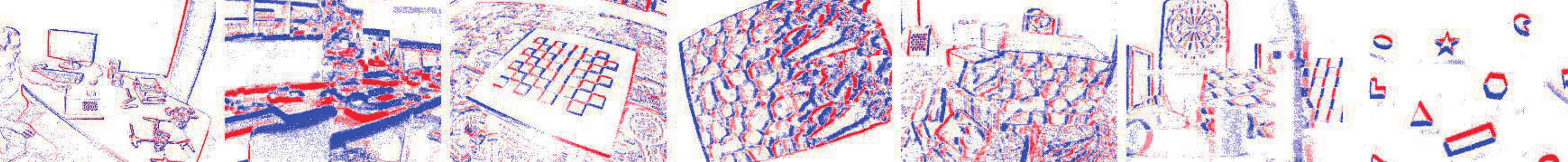}
		\caption{events.}
		\label{fig:eventsCnnMinimalEvents}
	\end{subfigure}
	\vfill
	\begin{subfigure}[b]{\linewidth}
		\includegraphics[width=\linewidth]{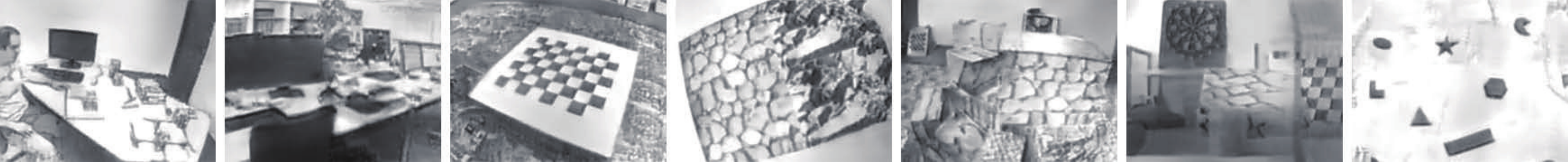}
		\caption{vid2e \cite{Rebecq2018, Gehrig2020}.}
		\label{fig:eventsCnnMinimalVid2e}
	\end{subfigure}
	\begin{subfigure}[b]{\linewidth}
		\includegraphics[width=\linewidth]{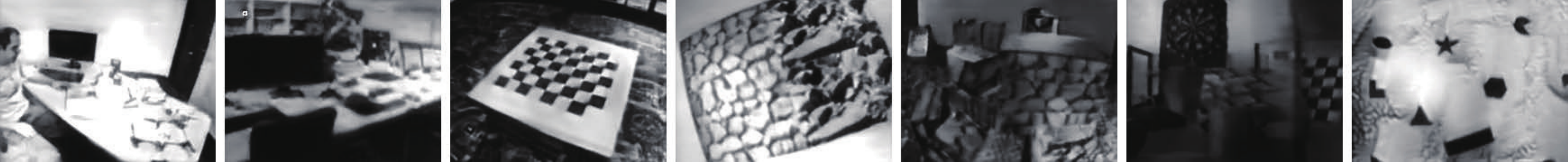}
		\caption{v2e \cite{Hu2021}.}
		\label{fig:eventsCnnMinimalV2e}
	\end{subfigure}
	\begin{subfigure}[b]{\linewidth}
		\includegraphics[width=\linewidth]{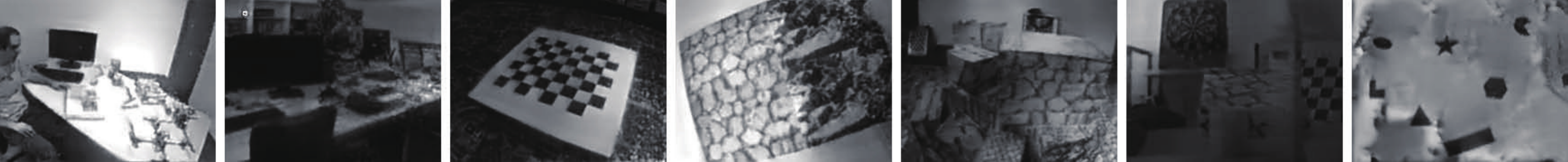}
		\caption{v2ce \cite{Zhang2024}.}
		\label{fig:eventsCnnMinimalV2ce}
	\end{subfigure}
	\begin{subfigure}[b]{\linewidth}
		\includegraphics[width=\linewidth]{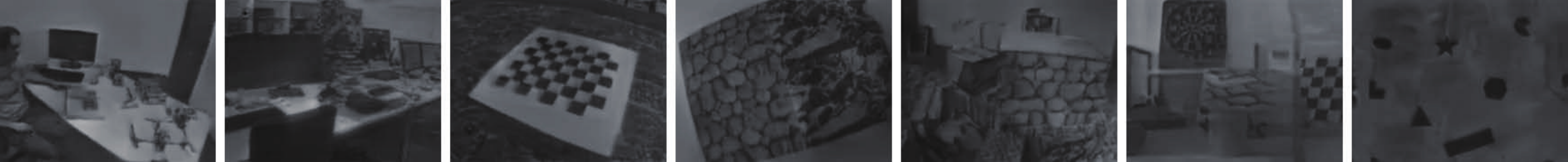}
		\caption{DVS-Voltmeter \cite{Lin2022}.}
		\label{fig:eventsCnnMinimalDVSVoltmeter}
	\end{subfigure}
	\begin{subfigure}[b]{\linewidth}
		\includegraphics[width=\linewidth]{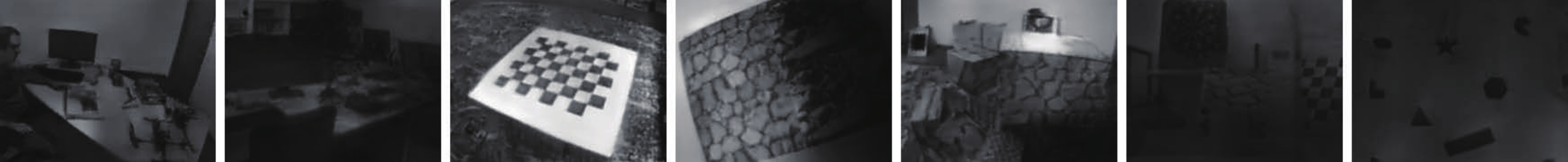}
		\caption{the proposed ADV2E.}
		\label{fig:eventsCnnMinimalADV2E}
	\end{subfigure}
	\begin{subfigure}[b]{\linewidth}
		\includegraphics[width=\linewidth]{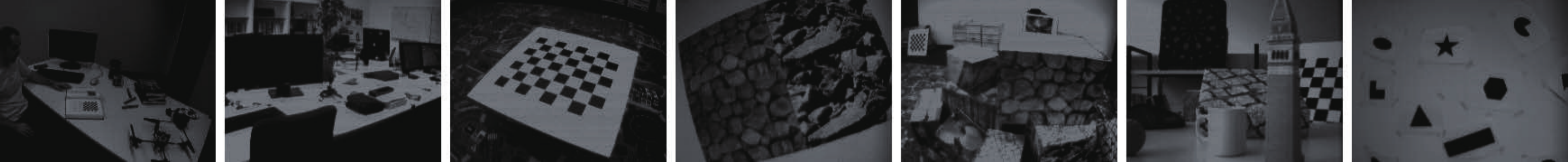}
		\caption{ground truth.}
		\label{fig:eventsCnnMinimalGT}
	\end{subfigure}
	\vspace{-6mm}
	\caption{Qualitative comparison of image reconstruction on datasets introduced in \cite{Mueggler2017}, with columns representing individual scenes from left to right as follows:  dynamic{\textunderscore}6dof, boxes{\textunderscore}6dof, poster{\textunderscore}6dof, shapes{\textunderscore}6dof, office{\textunderscore}zigzag,  slider{\textunderscore}depth, and calibration. From top to bottom, the rows show (a) events, segmentation results by (b) vid2e \cite{Rebecq2018, Gehrig2020}, (c) v2e \cite{Hu2021}, (d) v2ce \cite{Zhang2024}, (e) DVS-Voltmeter \cite{Lin2022}, (f) the proposed ADV2E, and (g) ground truth. Compared to other methods, the proposed approach generates events that train the network to reconstruct cleaner images with fewer artifacts and greater detail. For instance, in shapes{\textunderscore}6dof, the proposed method produces a background with minimal artifacts, and in office{\textunderscore}zigzag, it recovers the desk with the most details.}
	\label{fig:eventsCnnMinimal}
\end{figure*}

In Fig. \ref{fig:eventsCnnMinimal} and Table \ref{tab:eventsCnnMinimal}, we compare the reconstructed results both qualitatively and quantitatively. Among all methods, the proposed ADV2E achieves the best MSE and SSIM, and the second best LPIPS. 

Compared to other simulators, ADV2E generates events that train the network to produce reconstructed images closest to the ground truth. In the 1st column of Fig. \ref{fig:eventsCnnMinimal}, the items on the table are reconstructed with fine details and minimal artifacts through ADV2E. While v2e captures more details under the table, it introduces overexposure that diverges significantly from the ground truth. Similar improvements are seen in the rest columns of Fig. \ref{fig:eventsCnnMinimal}. In the 3rd column, the proposed method produces the fewest artifacts while preserving the most details. Other methods, such as v2e and DVS-Voltmeter, introduce noticeable artifacts, while vid2e and v2ce obscure fine details.

The ADV2E simulator accurately models the fundamental analogue behaviors of the DVS circuit, particularly the low-pass filter. With a cutoff frequency that adjusts instantaneously to brightness, the edges between light and dark areas generate events based on varying cutoff frequencies. Since ADV2E adapts to these frequency changes immediately, the simulated events are more realistic, resulting in reconstructions that are closer to the ground truth.

\section{Conclusion}

This work introduces ADV2E, an innovative event simulator designed to generate synthetic events from existing APS frames. The proposed method leverages the core analogue behaviors of the DVS circuit, particularly the low-pass filtering effect of analogue components. By incorporating these behaviors, ADV2E effectively bridges the long-standing gap between analogue pixel circuits and discrete video frames.

Compared to existing methods, ADV2E generates more realistic events, especially in high-contrast scenes. Its effectiveness is validated on two event-based tasks: semantic segmentation and image reconstruction. ADV2E achieves the highest mIoU and second-highest accuracy in semantic segmentation, as well as the best MSE , the second-best LPIPS, and the best SSIM scores in image reconstruction.

{
	\small
	\bibliographystyle{ieeenat_fullname}
	\bibliography{main}
}


\end{document}